\documentclass{article}
\usepackage[latin1]{inputenc}
\usepackage[sc]{mathpazo}
\usepackage[T1]{fontenc}
\usepackage{fullpage}
\usepackage{amsmath}
\usepackage{amsfonts}
\usepackage{amssymb}
\usepackage{graphicx}
\usepackage{verbatim}
\usepackage{hyperref}
\usepackage{authblk}
\usepackage[ruled,vlined]{algorithm2e}
\usepackage{mdwlist}
\usepackage{multirow}
\usepackage{multicol}

\hypersetup{
    colorlinks,
    citecolor=blue,
    filecolor=black,
    linkcolor=blue,
    urlcolor=blue,
    pdfauthor={Liangjie Hong},
    pdfsubject={General Writings},
    pdftitle={Tutorial on Probablistic Latent Semantic Analysis},
    pdfkeywords={machine learning, information retrieval, social media, social networks, data mining, web mining, recommendation, recommender systems, personalization, collaborative filtering}    
}

\title{Probabilistic Latent Semantic Analysis}
\author{Liangjie Hong}
\affil{Department of Computer Science and Engineering}
\affil{Lehigh University}
\begin{document}
\maketitle

\section{Some History}
Historically, many believe that these three papers \cite{Hofmann1999,Hofmann1999a,Hofmann2001} established the techniques of Probabilistic Latent Semantic Analysis or PLSA for short. However, there also exists one variant of the model in \cite{Hofmann1999b} and indeed all these models were originally discussed in an earlier technical report \cite{Hofmann1998}. In \cite{Chien2008}, the authors extended MLE-style estimation of PLSA to MAP-style estimations. A hierarchical extension was proposed in \cite{Hofmann1999c}. In \cite{Ding2008}, the authors showed the equivalent between PLSA and another popular method, non-negative matrix factorization. A high order of proof was shown in \cite{Peng2009}. The equivalent between PLSA and LDA was shown in \cite{Girolami2003}. More recently, a new MAP estimation algorithm is proposed in \cite{Taddy2012}.
\section{A Modern View of PLSA}
In order to better understand the intuition behind the model, we need to make some assumptions. First, we assume a topic $\phi_{k}$ is a distribution over a fixed size of vocabulary $V$. In the original PLSA model, this distribution is not explicitly specified but the form is in Multinomial distribution. Thus, $\phi_{k}$ is essentially a vector that each element $\phi_{(k,w)}$ represents the probability that term $w$ is chosen by topic $k$, namely:
\begin{equation}
p(w|k) = \phi_{(k,w)}
\end{equation}and note $\sum_{w}^{V} \phi_{(k,w)} = 1$. Secondly, we also assume that a document consists of multiple topics. Therefore, there is a distribution $\theta_{d}$ over a fixed number of topics $T$ for each document $d$. Similarly, original PLSA model does not have the explicit specification of this distribution but it is indeed a Multinomial distribution where each element $\theta_{(d,k)}$ in the vector $\theta_{d}$ represents the probability that topic $k$ appears in document $d$, namely:
\begin{equation}
p(k|d) = \theta_{(d,k)}
\end{equation}and also $\sum_{k}^{T} \theta_{(d,k)}=1$. This is the prerequisite of the model.

PLSA can be considered as a generative model, although it is not strictly the case \cite{Blei2003}. Before we start, there is one subtle issue needs to be pointed out. That is the difference between a term $w$ in the vocabulary $V$ and a token position $d_{i}$ in a document $d$. Terms in the vocabulary are distinct, meaning that all the terms differ from each other. Token positions are the places where terms are realized. Therefore, a term could appear multiple times in a same document $d$ in different token positions.

Imagine someone wants to write a document, he needs to decide which term to choose for each token position in a document $d$. For $i$-th position, he first decides which topic he wants to write, according to the distribution $\theta_{d}$. In this step, he essentially flips a $T$- side dice since $\theta_{d}$ is a Multinomial distribution. Once the outcome of decision is made, suppose it is topic $k$, he then chooses a term, according to the distribution $\phi_{k}$. Similarly, a $V$-side dice is flipped. This two step generation process is repeated for all token positions and for all documents in the dataset.

The generation process can be summarized as follows:
\begin{itemize}
\item For each document $d$
		\begin{itemize}
		\item For each token position $i$ \\
			  Choose a topic $z$ $\sim$ Multinomial($\theta_{d}$) \\
			  Choose a term $w$ $\sim$ Multinomial($\phi_{z}$)
		\end{itemize}
\end{itemize}and we can write the probability a term $w$ appearing at token position $i$ in document $d$ as follows:
\begin{equation}
p(d_{i} = w | \Phi, \theta_{d}) = \sum_{z=k}^{T} \phi_{(z,w)} \theta_{(d,z)}
\end{equation}and the joint likelihood of the whole dataset $\mathcal{W}$ is:
\begin{eqnarray}
p(\mathcal{W}|\Phi, \Theta) &=& \prod_{d}^{D} \prod_{i}^{N_{d}} \sum_{z=k}^{T} \phi_{(z,w)} \theta_{(d,z)} \nonumber \\
	 &=& \prod_{d}^{D} \prod_{w}^{V} \Bigr( \sum_{z=k}^{T} \phi_{(z,w)} \theta_{(d,z)} \Bigl)^{n(d,w)}
\end{eqnarray}where $n(d,w)$ is the number of times term $w$ appearing in document $d$.

In the formalism above, the likelihood depends on parameters $\Phi$ and $\Theta$, which needs to be estimated from data. Here, we wish to obtain the parameters that can maximize the above likelihood. Therefore, we have:
\begin{eqnarray}
\arg\max_{\Phi, \Theta} \Bigr[ \log p(\mathcal{W}|\Phi, \Theta) + \sum_{d}^{D} \lambda_{d} ( 1 - \sum_{z}^{T}  \theta_{(d,z)}) + \sum_{z}^{T}  \sigma_{k} ( 1 - \sum_{w}^{V} \phi_{(z,w)}) \Bigl]
\end{eqnarray}where the second and the third part of the equation is Lagrange Multipliers to guarantee Multinomial parameters in range $[0,1]$.

It is difficult to directly optimize the above equation due to the $\log$ sign is out of a summation. EM (Expectation Maximization) \cite{Dempster1977} algorithm is employed here to estimate these parameters. The key assumption to apply EM algorithm is that we know for each token position which topic is chosen from. In other words, for each token position, we know $z$ value. Note, we just \textit{pretend} we know these values. We denote $R_{w_{di}}$ to represent which $z$ is chosen for token position $di$ in document $d$. Thus, $R_{w_{di}}$ is a $T$ dimensional vector where $\sum_{k} R_{(w_{di},k)} = 1$. This also indicates that each $R_{w_{di}}$ is in fact a valid distribution and $\mathbf{R}$ is a matrix where each row entry is a $R_{w_{di}}$. We plug all these hidden variables into the likelihood function:
\begin{eqnarray}
\mathcal{L} = \log p(\mathcal{W}|\mathbf{R}, \Phi, \Theta) = \sum_{d}^{D} \sum_{di}^{N_{d}} \sum_{z}^{T} R_{(w_{di},z)} \Bigr( \log \phi_{(z,w_{di})} +\log \theta_{(d,z)} \Bigl)
\end{eqnarray}and our new objective function is as follows:
\begin{eqnarray}
\arg\max_{\Phi, \Theta} \Lambda = \Bigr[ \log p(\mathcal{W}|\mathbf{R},\Phi, \Theta) + \sum_{d}^{D} \lambda_{d} ( 1 - \sum_{z}^{T} \theta_{(d,z)}) + \sum_{z}^{T} \sigma_{k} ( 1 - \sum_{w}^{V} \phi_{(z,w)}) \Bigl]
\end{eqnarray}For a standard E-step in EM algorithm, we compute the posterior distribution of hidden variables, given the data and the current values of parameters:
{\large{\begin{eqnarray}\label{eq:lda_e_step}
< R_{(w_{di},k)} > &=& p(R_{(w_{di},k)} = 1 | \mathcal{W}, \Theta, \Phi) \nonumber \\
&=& \frac{p(\mathcal{W},R_{(w_{di},k)}=1|\Theta, \Phi)}{\sum_{k}^{T} p(\mathcal{W},R_{(w_{di},k)}=1|\Theta, \Phi)} \nonumber \\
&=& \frac{p(w_{di},R_{(w_{di},k)}=1|\theta_{d}, \Phi)}{\sum_{k}^{T} p(w_{di},R_{(w_{di},k)}=1|\theta_{d}, \Phi)} \nonumber \\
&=& \frac{p(w_{di}|\phi_{(k,w_{di})})p(k|\theta_{d})}{\sum_{k}^{T} p(w_{di}|\phi_{(k,w_{di})})p(k|\theta_{d})} \nonumber \\
&=& \frac{\phi_{(k,w_{di})} \theta_{(d,k)}}{\sum_{k}^{T} \phi_{(k,w_{di})} \theta_{(d,k)}}
\end{eqnarray}}}In M-step, we obtain the new optimal values for parameters given the current settings of hidden variables. For $\theta_{d}$, we have:
\begin{eqnarray}
\frac{\partial \Lambda}{\partial \theta_{(d,z)}} &=& \sum_{di}^{N_{d}} \frac{<R_{(w_{di},z)}>}{\theta_{(d,z)}} - \lambda_{d} = 0  \nonumber \\
\frac{\partial \Lambda}{\partial \lambda_{d}}    &=& 1 - \sum_{z}^{T} \theta_{(d,z)} = 0 \nonumber \\
\end{eqnarray}Solving the above two equations, we obtain:
\begin{equation}\label{eq:lda_m_step_theta}
\theta_{(d,z)} = \frac{\sum_{di} <R_{(w_{di},z)}>}{N_{d}}
\end{equation}Similarly, for $\phi_{z}$, we have:
\begin{eqnarray}
\frac{\partial \Lambda}{\partial \phi_{(z,w)}} &=& \sum_{d}^{D} \sum_{di}^{N_{d}} \frac{<R_{(w_{di},z)>}\mathbb{I}(w_{di}=w)}{\theta_{(z,w)}} - \sigma_{z} = 0 \nonumber \\
\frac{\partial \Lambda}{\partial \sigma_{z}} &=& 1 - \sum_{w}^{V} \phi_{(z,w)} = 0 \nonumber \\
\end{eqnarray}Solving the above two equations, we obtain:
\begin{equation}\label{eq:lda_m_step_phi}
\phi_{(z,w)} = \frac{\sum_{d}^{D} \sum_{di}^{N_{d}} <R_{(w_{di},z)>}\mathbb{I}(w_{di}=w)}{\sum_{w'}^{V} \sum_{d}^{D} \sum_{di}^{N_{d}} <R_{(w_{di},z)>}\mathbb{I}(w_{di}=w')}
\end{equation}Note, we can simplify the notation of EM step. Notice that for all token positions of a same term $w$ in a same document $d$, E-step is essentially same and therefore simplified E-step is:
\begin{eqnarray}\label{eq:lda_simplified_e_step}
<R_{(w,k)}^{(d)}> = \frac{\phi_{(k,w)} \theta_{(d,k)}}{\sum_{k}^{T} \phi_{(k,w)} \theta_{(d,k)}}
\end{eqnarray} and simplified M-step is:
\begin{eqnarray}
\theta_{(d,k)} &=& \frac{\sum_{w}^{V} n(d,w) <R_{(w,k)}^{(d)}>}{N_{d}} \nonumber \\
\phi_{(k,w)} &=& \frac{\sum_{d}^{D} n(d,w) <R_{(w,k)}^{(d)}>}{\sum_{w'}^{V} \sum_{d}^{D} n(d,w') <R_{(w',k)}^{(d)}>}
\end{eqnarray}
\section{Discussion on EM Algorithm}
In the above discussion, there is one subtle detail that needs more space to be clarified. We introduced $R_{(w_{di},k)}$ as indicator variables to indicate which topic is chosen for token position $di$. Although it satisfies $\sum_{k} R_{(w_{di},k)} = 1$, this vector essentially only has one element equal to $1$. However, when we calculate E-step of the inference algorithm, we calculate $<R_{(w_{di},k)}>$, the posterior distribution of hidden variables, given the data and current settings of parameters. Here, $<R_{(w_{di},k)}>$ is a distribution and it has probabilities in each element of the vector but still satisfies $\sum_{k} <R_{(w_{di},k)}> = 1$. What really leads to this difference?

We re-write the log likelihood of one token position after we introduce the indicator variables as follows:
\begin{equation}
\log \sum_{k}^{T} R_{(w_{di},k)} \Bigr( \phi_{(k,w_{di})} \theta_{(d,k)} \Bigl)
\end{equation}We introduce an auxiliary distribution $q(R_{(w_{di},k)}) = q(R_{(w_{di},k)} = 1)$ and therefore $\sum_{k} q(R_{(w_{di},k)}) = 1$. Plug this auxiliary distribution into the above log likelihood, we obtain:
\begin{equation}
\log \sum_{k}^{T} \frac{R_{(w_{di},k)} \Bigr( \phi_{(k,w_{di})} \theta_{(d,k)} \Bigl)}{q(R_{(w_{di},k)})} q(R_{(w_{di},k)}) =
\log \mathbb{E}_{q} \Bigr[ \frac{R_{(w_{di},k)} \Bigr( \phi_{(k,w_{di})} \theta_{(d,k)} \Bigl)}{q(R_{(w_{di},k)})} \Bigl]
\end{equation}By using Jensen's Inequality, we can move the log sign into the expectation and make a lower bound of our original log likelihood:
\begin{eqnarray}
\log \mathbb{E}_{q} \Bigr[ \frac{R_{(w_{di},k)} \Bigr( \phi_{(k,w_{di})} \theta_{(d,k)} \Bigl)}{q(R_{(w_{di},k)})} \Bigl] &\geq & \mathbb{E}_{q} \Bigr[ \log \frac{R_{(w_{di},k)} \Bigr( \phi_{(k,w_{di})} \theta_{(d,k)} \Bigl)}{q(R_{(w_{di},k)})} \Bigl] \nonumber \\
&\geq & \mathbb{E}_{q} \Bigr[ \log \Bigr( R_{(w_{di},k)} \Bigr( \phi_{(k,w_{di})} \theta_{(d,k)} \Bigl) \Bigl) - \log q(R_{(w_{di},k)}) \Bigl]
\end{eqnarray}Now, our goal is clear. Since it is hard to directly optimize the left hand side, we need to maximize the lower bound, right hand side, as much as possible:
\begin{eqnarray}
\sum_{k} q(R_{(w_{di},k)}) \log \Bigr( R_{(w_{di},k)} \Bigr( \phi_{(k,w_{di})} \theta_{(d,k)} \Bigl) \Bigl) - \sum_{k} q(R_{(w_{di},k)}) \log q(R_{(w_{di},k)})
 + \lambda \Bigr( 1 - \sum_{k} q(R_{(w_{di},k)}) \Bigl)
\end{eqnarray}Taking the derivatives respect to $q(R_{(w_{di},k)})$ and setting to $0$, we have:
\begin{equation}
\log \Bigr( R_{(w_{di},k)} \Bigr( \phi_{(k,w_{di})} \theta_{(d,k)} \Bigl) \Bigl) - \log q(R_{(w_{di},k)}) - 1 - \lambda = 0
\end{equation}Solving this, we obtain:
\begin{equation}
q(R_{(w_{di},k)}) = \frac{\phi_{(k,w_{di})} \theta_{(d,k)}}{\sum_{k}^{T} \phi_{(k,w_{di})} \theta_{(d,k)}}
\end{equation}It is exactly E-step we obtained in the previous section. Note, $q(R_{(w_{di},k)})$ is indeed $<R_{(w_{di},k)}>$ and we understand that EM algorithm here in a lower bound maximization process.
\section{Original Formalism of PLSA}
In original proposed PLSA by Thomas Hofmann \cite{Hofmann1999,Hofmann1999a,Hofmann2001}, there are two ways to formulate PLSA. They are equivalent but may lead to different inference process.
\begin{equation}
P(d,w) = P(d) \sum_{z} P(w|z)P(z|d)
\end{equation}
\begin{equation}
P(d,w) = \sum_{z} P(w|z)P(d|z)P(z)
\end{equation}
Let's see why these two equations are equivalent by using Bayes rule.
\begin{displaymath}
\begin{aligned}
P(z|d) &= \frac{P(d|z)P(z)}{P(d)} \\
P(z|d)P(d) &=  P(d|z)P(z)\\
P(w|z)P(z|d)P(d) &=  P(w|z)P(d|z)P(z)\\
P(d) \sum_{z} P(w|z)P(z|d) &= \sum_{z} P(w|z)P(d|z)P(z)
\end{aligned}
\end{displaymath}
The whole data set is generated as (we assume that all words are generated independently):
\begin{equation}
D = \prod_{d} \prod_{w} P(d,w)^{n(d,w)}
\end{equation}
The Log-likelihood of the whole data set for (1) and (2) are:
\begin{equation}\label{eq:plsa_loglikelihood_1}
L_{1} = \sum_{d} \sum_{w} n(d,w) \log [ P(d) \sum_{z} P(w|z)P(z|d) ]
\end{equation}
\begin{equation}\label{eq:plsa_loglikelihood_2}
L_{2} = \sum_{d} \sum_{w} n(d,w) \log [ \sum_{z} P(w|z)P(d|z)P(z) ]
\end{equation}
\section{EM}
For Equation \ref{eq:plsa_loglikelihood_1} and Equation \ref{eq:plsa_loglikelihood_2}, the optimization is hard due to the log of sum. Therefore, an algorithm called Expectation-Maximization is usually employed. Before we introduce anything about EM, please note that EM is only guarantee to find a local optimum, although it may be a global one.

First, we see how EM works in general. As we shown for PLSA, we usually want to estimate the likelihood of data, namely $P(X|\theta)$, given the paramter $\theta$. The easiest way is to obtain a maximum likelihood estimator by maximizing $P(X|\theta)$. However, sometimes, we also want to include some hidden variables which are usually useful for our task. Therefore, what we really want to maximize is $P(X|\theta)=\sum_{z}P(X|z,\theta)P(z|\theta)$, the complete likelihood. Now, our attention becomes to this complete likelihood. Again, directly maximizing this likelihood is usually difficult. What we would like to show here is to obtain a lower bound of the likelihood and maximize this lower bound.

We need Jensen's Inequality to help us obtain this lower bound. For any convex function $f(x)$, Jensen's Inequality states that :
\begin{align}
\lambda f(x) + (1-\lambda) f(y) \geq f(\lambda x + (1-\lambda) y)
\end{align} Thus, it is not difficult to show that :
\begin{align}
\mathbb{E}[f(x)] = \sum_{x} P(x) f(x) \geq f(\sum_{x} P(x) x) = f(\mathbb{E}[x])
\end{align} For concave functions (like logarithm), Jensen's Inequality should be used reversely as:
\begin{align}
\mathbb{E}[f(x)] \leq f(\mathbb{E}[x])
\end{align} Back to our complete likelihood, we can obtain the following conclusion by using concave version of Jensen's Inequality :
\begin{align}
\log \sum_{z}P(X|z,\theta)P(z|\theta)&= \log \sum_{z}P(X|z,\theta)P(z|\theta)\frac{q(z)}{q(z)} \\
 &= \log \mathbb{E}_{q} \Bigr[ \frac{P(X|z,\theta)P(z|\theta)}{q(z)} \Bigl] \\
 & \geq \mathbb{E}_{q} \Bigr[ \log \frac{P(X|z,\theta)P(z|\theta)}{q(z)} \Bigl]
\end{align}where $\mathbb{E}_{q}$ is expectation with respect to $q(z)$. Therefore, we obtained a lower bound of complete likelihood and we want to maximize it as tight as possible. EM is an algorithm that maximize this lower bound through an iterative fashion. Usually, EM first would fix current $\theta$ value and maximize $q(z)$ and then use the new $q(z)$ value to obtain a new guess on $\theta$, which is essentially a two stage maximization process. The first step can be shown as follows:
\begin{align*}
\mathbb{E}_{q} \Bigr[ \log \frac{P(X|z,\theta)P(z|\theta)}{q(z)} \Bigl] &= \sum_{z} q(z) \log \frac{P(X|z,\theta)P(z|\theta)}{q(z)}\\
&= \sum_{z} q(z) \log \frac{P(z \, | \, X,\theta)P(X \, | \, \theta)}{q(z)}\\
&= \sum_{z} q(z) \log P(X \, | \, \theta) + \sum_{z} q(z) \log \frac{P(z|X,\theta)}{q(z)}\\
&= \log P(X \, | \, \theta) - \sum_{z} q(z) \log \frac{q(z)}{P(z|X,\theta)}\\
&= \log P(X \, | \, \theta) - \mathbb{E}_{q} \Bigr[ \log  \frac{q(z)}{P(z|X,\theta)} \Bigl]\\
&= \log P(X \, | \, \theta) - KL \Bigr( q(z) \, || \,P(z|X,\theta) \Bigl)
\end{align*}The first term does not contain $z$. Therefore, in order to maximize the whole equation, we need to minimize KL divergence between $q(z)$ and $P(z|X,\theta)$, which eventually leads to the optimum solution of $q(z) = P(z|X,\theta)$. So, usually for E-step, we use current guess of $\theta$ to calculate the posterior distribution of hidden variable as the new update score. For M-step, it is problem-dependent. We will see how to do that in later discussions.

We also show another explanation of EM in terms of optimizing a so-called Q function. We devise the data generation process as $P(X|\theta)=P(X,H|\theta)=P(H|X,\theta)P(X|\theta)$. Therefore, the complete likelihood is modified as:
\begin{align*}
L_{c}(\theta) = \log P(X,H|\theta) = \log P(X|\theta) + \log P(H|X,\theta) = L(\theta) + \log P(H|X,\theta)
\end{align*}Think about how to maximize $L_{c}(\theta)$. Instead of directly maximizing it, we can iteratively maximize $L_{c}(\theta^{(n+1)})-L_{c}(\theta^{(n)})$ as :
\begin{align*}
L(\theta) - L(\theta^{(n)}) &= L_{c}(\theta) - \log P(H|X,\theta) - L_{c}(\theta^{(n)}) + \log P(H|X,\theta^{(n)})\\
							&= L_{c}(\theta) - L_{c}(\theta^{(n)}) + \log \frac{P(H|X,\theta^{(n)})}{P(H|X,\theta)}
\end{align*}Now take the expectation of this equation, we have:
\begin{align*}
L(\theta) - L(\theta^{(n)}) = \sum_{H} L_{c}(\theta)P(H|X,\theta^{(n)}) - \sum_{H} L_{c}(\theta^{(n)})P(H|X,\theta^{(n)}) + \sum_{H} P(H|X,\theta^{(n)})\log \frac{P(H|X,\theta^{(n)})}{P(H|X,\theta)}
\end{align*}The last term is always non-negative since it can be recognized as the KL-divergence of $P(H|X,\theta^{(n)}$ and $P(H|X,\theta)$. Therefore, we obtain a lower bound of Likelihood : \begin{align*}
L(\theta) \geq \sum_{H} L_{c}(\theta)P(H|X,\theta^{(n)}) + L(\theta^{(n)}) - \sum_{H} L_{c}(\theta^{(n)})P(H|X,\theta^{(n)})
\end{align*}The last two terms can be treated as constants as they do not contain the variable $\theta$, so the lower bound is essentially the first term, which is also sometimes called as ``Q-function''. \begin{equation}
Q(\theta;\theta^{(n)}) = E(L_{c}(\theta)) = \sum_{H} L_{c}(\theta) P (H|X,\theta^{(n)})
\end{equation}
\subsection{EM of Formulation 1}
In case of Formulation 1, let us introduce hidden variables $R(z,w,d)$ to indicate which hidden topic $z$ is selected to generated $w$ in $d$ where $\sum_{z} R(z,w,d) = 1$. Therefore, the complete likelihood can be formulated as:
\begin{align*}
L_{c1} &= \sum_{d} \sum_{w} n(d,w) \sum_{z} R(z,w,d) \log \Bigr[ P(d) P(w \, | \, z)P(z \, | \, d) \Bigl] \\
	   &= \sum_{d} \sum_{w} n(d,w) \sum_{z} R(z,w,d) \Bigr[ \log P(d) + \log P(w|z) + \log P(z \, | \, d) \Bigl]
\end{align*}From the equation above, we can write our Q-function for the complete likelihood $\mathbb{E}[L_{c1}]$:
\begin{align*}
\mathbb{E}[L_{c1}] & = \sum_{d} \sum_{w} n(d,w) \sum_{z} P(z|w,d) \Bigr[ \log P(d) + \log P(w\, |\, z) + \log P(z\, | \, d) \Bigl]
\end{align*}For E-step, we obtain the posterior probability for latent variables:
\begin{align*}
P(z|w,d) & = \frac{P(w,z,d)}{P(w,d)}\\
		 & = \frac{P(w|z)P(z|d)P(d)}{\sum_{z} P(w|z)P(z|d)P(d)}\\
		 & = \frac{P(w|z)P(z|d)}{\sum_{z} P(w|z)P(z|d)}
\end{align*}For M-step, we need to maximize Q-function, which needs to be incorporated with other constraints:
\begin{align*}
H & = \mathbb{E}[L_{c1}] + \alpha [1-\sum_{d} P(d) ] + \beta \sum_{z}[1- \sum_{w} P(w|z)]+  \gamma \sum_{d}[1- \sum_{z} P(z|d)]
\end{align*}where $\alpha$, $\beta$ and $\gamma$ are Lagrange Multipliers. We take all derivatives:
\begin{align*}
\frac{\partial H}{\partial P(d)} &= \sum_{w} \sum_{z} n(d,w) \frac{P(z|w,d)}{P(d)} - \alpha = 0\\
								 &\rightarrow \sum_{w} \sum_{z} n(d,w) P(z|w,d) - \alpha P(d) = 0\\
\frac{\partial H}{\partial P(w|z)} &= \sum_{d} n(d,w) \frac{P(z|w,d)}{P(w|z)} - \beta = 0\\
								 &\rightarrow \sum_{d} n(d,w) P(z|w,d) - \beta P(w|z) = 0\\
\frac{\partial H}{\partial P(z|d)} &= \sum_{w} n(d,w) \frac{P(z|w,d)}{P(z|d)} - \gamma = 0\\
								 &\rightarrow \sum_{w} n(d,w) P(z|w,d) - \gamma P(z|d) = 0
\end{align*}Therefore, we can easily obtain:
\begin{align}
P(d) &= \frac{\sum_{w} \sum_{z} n(d,w) P(z|w,d)}{\sum_{d} \sum_{w} \sum_{z} n(d,w) P(z|w,d)} \nonumber \\
	 &= \frac{n(d)}{\sum_{d} n(d)}\\
P(w|z) &= \frac{\sum_{d} n(d,w) P(z|w,d)}{\sum_{w} \sum_{d} n(d,w) P(z|w,d) } \\
P(z|d) &= \frac{\sum_{w} n(d,w) P(z|w,d)}{\sum_{z} \sum_{w} n(d,w) P(z|w,d) } \nonumber \\
	   &= \frac{\sum_{w} n(d,w) P(z|w,d)}{n(d)}
\end{align}
\subsection{EM of Formulation 2}
Use similar method to introduce hidden variables to indicate which $z$ is selected to generated $w$ and $d$ and we can have the following complete likelihood :
\begin{align*}
L_{c2} &= \sum_{d} \sum_{w} n(d,w) \sum_{z} R(z,w,d) \log [ P(z) P(w|z)P(d|z) ] \\
	   &= \sum_{d} \sum_{w} n(d,w) \sum_{z} R(z,w,d) \Bigr[ \log P(z) + \log P(w|z) + \log P(d|z) \Bigl]
\end{align*}Therefore, the Q-function $E[L_{c2}]$ would be :
\begin{align*}
\mathbb{E}[L_{c2}] & = \sum_{d} \sum_{w} n(d,w) \sum_{z} P(z|w,d) [ \log P(z) + \log P(w|z) + \log P(d|z) ]
\end{align*}For E-step, again, we obtain the posterior probability for latent variables:
\begin{align*}
P(z|w,d) & = \frac{P(w,z,d)}{P(w,d)}\\
		 & = \frac{P(w|z)P(d|z)P(z)}{\sum_{z} P(w|z)P(d|z)P(z)}
\end{align*}
For M-step, we maximize the constraint version of Q-function:
\begin{align*}
H & = \mathbb{E}[L_{c2}] + \alpha [1-\sum_{z} P(z) ] + \beta \sum_{z}[1- \sum_{w} P(w|z)]+  \gamma \sum_{z} [1- \sum_{d} P(d|z)]
\end{align*}where $\alpha$, $\beta$ and $\gamma$ are Lagrange Multipliers. We take all derivatives:
\begin{align*}
\frac{\partial H}{\partial P(z)} &= \sum_{d} \sum_{w} n(d,w) \frac{P(z|w,d)}{P(z)} - \alpha = 0\\
								 &\rightarrow \sum_{d} \sum_{w} n(d,w) P(z|w,d) - \alpha P(z)= 0\\
\frac{\partial H}{\partial P(w|z)} &= \sum_{d} n(d,w) \frac{P(z|w,d)}{P(w|z)} - \beta = 0\\
								 &\rightarrow \sum_{d} n(d,w) P(z|w,d) - \beta P(w|z) = 0\\
\frac{\partial H}{\partial P(d|z)} &= \sum_{w} n(d,w) \frac{P(z|w,d)}{P(d|z)} - \gamma = 0\\
								 &\rightarrow \sum_{w} n(d,w) P(z|w,d) - \gamma P(d|z) = 0
\end{align*}Therefore, we can easily obtain:
\begin{align}
P(z) &= \frac{\sum_{d} \sum_{w} n(d,w) P(z|w,d)}{\sum_{d} \sum_{w} \sum_{z} n(d,w) P(z|w,d)} \nonumber \\
	 &= \frac{\sum_{d} \sum_{w} n(d,w) P(z|w,d)}{\sum_{d} \sum_{w} n(d,w)} \\
P(w|z) &= \frac{\sum_{d} n(d,w) P(z|w,d)}{\sum_{w} \sum_{d} n(d,w) P(z|w,d) } \\
P(d|z) &= \frac{\sum_{w} n(d,w) P(z|w,d)}{\sum_{d} \sum_{w} n(d,w) P(z|w,d) }
\end{align}
\section{Incorporating Background Language Model}
Another PLSA model which incorporates background language model is usually formulated like this : \begin{equation}
P(d,w) = \lambda_{B} P (w|\theta_{B}) + (1-\lambda_{B}) \sum_{z} P(w|z)P(z|d)P(d)
\end{equation}
The log likelihood of Equation 7 is \begin{displaymath}
L = \sum_{d} \sum_{w} n(d,w) \log [ \lambda_{B} P (w|\theta_{B}) + (1-\lambda_{B}) \sum_{z} P(w|z)P(z|d)P(d) ]
\end{displaymath}
Let's again introduce a hidden variable $P(Z_{d,w})$ to indicate which component that the $w$ and $d$ are generated while $P(Z_{d,w} = \theta_{B})$ means that the word is generated by the background model and $P(Z_{d,w} = j)$ meaning the word is generated by the topic $z_{j}$. Thus, the complete log likelihood is : \begin{displaymath}
L_{c} = \sum_{d} \sum_{w} n(d,w) [ P(Z_{d,w} = \theta_{B}) \log (\lambda_{B} P (w|\theta_{B})) + \sum_{z} P(Z_{d,w} = z|Z_{d,w} \neq \theta_{B}) \log ((1-\lambda_{B})P(w|z)P(z|d)P(d))]
\end{displaymath}
The E-step is straightforward. Using Bayes Rule, we can obtain: \begin{displaymath}
\begin{aligned}
P(Z_{d,w} = \theta_{B} | d,w) &= \frac{P(w|\theta_{B},d)}{P(w,d)}\\
							  &= \frac{\lambda_{B}P(w|\theta_{B})}{\lambda_{B} P (w|\theta_{B}) + (1-\lambda_{B}) \sum_{z} P(w|z)P(z|d)P(d)}\\
 P(Z_{d,w} = z | d,w) &= \frac{P(w|z,d)}{P(w,d)}\\
							  &= \frac{P(w|z)P(z|d)P(d)}{\sum_{z} P(w|z)P(z|d)P(d)}\\
							  &= \frac{P(w|z)P(z|d)}{\sum_{z} P(w|z)P(z|d)}
\end{aligned}
\end{displaymath}
For M-step, we maximize the constraint version of Q-function:  \begin{displaymath}
\begin{aligned}
H & = E[L_{c}] + \beta [1- \sum_{w} P(w|z)]+  \gamma [1- \sum_{z} P(z|d)]\\
\end{aligned}
\end{displaymath}
and take all derivatives: \begin{displaymath}
\begin{aligned}
\frac{\partial H}{\partial P(w|z)} &= \sum_{d} n(d,w) \frac{P(Z_{d,w}=z)}{P(w|z)} - \beta = 0\\
								 &\rightarrow \sum_{d} n(d,w) P(Z_{d,w}=z) - \beta P(w|z) = 0\\
\frac{\partial H}{\partial P(z|d)} &= \sum_{w} n(d,w) \frac{P(Z_{d,w}=z)}{P(z|d)} - \gamma = 0\\
								 &\rightarrow \sum_{w} n(d,w) P(Z_{d,w}=z) - \gamma P(z|d) = 0\\
\end{aligned}
\end{displaymath}
Therefore, we can easily obtain: \begin{displaymath}
\begin{aligned}
P(w|z) &= \frac{\sum_{d} n(d,w) (1-P(Z_{d,w} = \theta_{B} | d,w))P(Z_{d,w}=z)}{\sum_{w} \sum_{d} n(d,w)(1-P(Z_{d,w} = \theta_{B} | d,w)) P(Z_{d,w}=z) } \\
P(z|d) &= \frac{\sum_{w} n(d,w) (1-P(Z_{d,w} = \theta_{B} | d,w))P(Z_{d,w}=z)}{\sum_{z} \sum_{w} n(d,w)(1-P(Z_{d,w} = \theta_{B} | d,w))P(Z_{d,w}=z) } \\
\end{aligned}
\end{displaymath}
Note, $P(w|\theta_{B})$ is only sampled once by using the equation: \begin{displaymath}
P(w|\theta_{B}) = \frac{\sum_{d} n(d,w)}{\sum_{w} \sum_{d} n(d,w)}
\end{displaymath}
If we change to the PLSA Formulation 2, we will get the following E steps: \begin{displaymath}
\begin{aligned}
P(Z_{d,w} = \theta_{B} | d,w) &= \frac{P(w|\theta_{B},d)}{P(w,d)}\\
							  &= \frac{\lambda_{B}P(w|\theta_{B})}{\lambda_{B} P (w|\theta_{B}) + (1-\lambda_{B}) \sum_{z} P(w|z)P(d|z)P(z)}\\
 P(Z_{d,w} = z | d,w) &= \frac{P(w|z,d)}{P(w,d)}\\
							  &= \frac{P(w|z)P(d|z)P(z)}{\sum_{z} P(w|z)P(d|z)P(z)}\\
\end{aligned}
\end{displaymath}
and corresponding M steps:\begin{displaymath}
\begin{aligned}
P(w|z) &= \frac{\sum_{d} n(d,w) (1-P(Z_{d,w} = \theta_{B} | d,w))P(Z_{d,w}=z)}{\sum_{w} \sum_{d} n(d,w)(1-P(Z_{d,w} = \theta_{B} | d,w)) P(Z_{d,w}=z) } \\
P(d|z) &= \frac{\sum_{w} n(d,w) (1-P(Z_{d,w} = \theta_{B} | d,w))P(Z_{d,w}=z)}{\sum_{d} \sum_{w} n(d,w)(1-P(Z_{d,w} = \theta_{B} | d,w))P(Z_{d,w}=z) } \\
P(z) &= \frac{\sum_{d} \sum_{w} n(d,w) (1-P(Z_{d,w} = \theta_{B} | d,w))P(Z_{d,w}=z))}{\sum_{d} \sum_{w} \sum_{z} n(d,w) (1-P(Z_{d,w} = \theta_{B} | d,w))P(Z_{d,w}=z)} \\
\end{aligned}
\end{displaymath}

\section{Acknowledgement}
I would like to thank \href{http://www.lehigh.edu/~xiz307/}{Xingjian Zhang} (Lehigh Univ.), \href{http://www.lehigh.edu/~yil712/}{Yi Luo} (Lehigh Univ.) and \href{mailto:luming1206@gmail.com}{Ming Lu} for pointing out errors and typos in the manuscript.

\bibliographystyle{abbrv}

\end{document}